\documentclass[10pt,twocolumn,letterpaper]{article}

\usepackage{iccv}
\usepackage{times}
\usepackage{epsfig}
\usepackage{graphicx}
\usepackage{amsmath}
\usepackage{amssymb}
\usepackage{booktabs}
\usepackage{multirow}
\usepackage{siunitx}
\usepackage[accsupp]{axessibility}  
\usepackage{comment}
\usepackage[usenames,dvipsnames,table]{xcolor}
\usepackage{tabularray}

\usepackage{enumitem}
\setlist{itemsep=0.1cm,topsep=0pt,parsep=0pt,partopsep=0pt,leftmargin=0.3cm}

\renewcommand{\paragraph}[1]{\vspace{2pt plus 1pt minus 1pt}\noindent{\bf #1}\;}

\usepackage[pagebackref=true,breaklinks=true,letterpaper=true,colorlinks,bookmarks=false]{hyperref}

\iccvfinalcopy 

\ificcvfinal\fi

\begin{document}

\title{CASSPR: Cross Attention Single Scan Place Recognition}


\author{
	\small
	\begin{tabular}{c c c c c c c }                                            
		Yan Xia$^{*1,2,3 \footnotemark[2]}$ &
		Mariia Gladkova$^{*1,5}$ &
		Rui Wang$^4$ &
		Qianyun Li$^1$ &
		Uwe Stilla$^1$ &
		João F. Henriques$^3$ &
		Daniel Cremers$^{1,2,3,5}$ \\                                        
		\multicolumn{7}{c}{\shortstack{$^1$Technical University of Munich $^2$Munich Center for Machine Learning (MCML) \\ $^3$ Visual Geometry Group, University of Oxford $^4$ Microsoft Zurich $^5$ Munich Data Science Institute}} \\                                                
		\multicolumn{7}{c}{\{yan.xia, mariia.gladkova, stilla, cremers\}@tum.de, qianyunli0@outlook.com, joao@robots.ox.ac.uk } 
	\end{tabular}                                                                       
}

\maketitle
\let\oldthefootnote\thefootnote
\renewcommand{\thefootnote}{\fnsymbol{footnote}} 
\footnotetext[2]{Corresponding author. * Equal contribution.} 
\let\thefootnote\oldthefootnote
\ificcvfinal\thispagestyle{empty}\fi

\begin{abstract}
Place recognition based on point clouds (LiDAR) is an important component for autonomous robots or self-driving vehicles.  
Current SOTA performance is achieved on accumulated LiDAR submaps using either point-based or voxel-based structures. While voxel-based approaches nicely integrate spatial context across multiple scales, they do not exhibit the local precision of point-based methods.  As a result, existing methods struggle with fine-grained matching of subtle geometric features in sparse single-shot LiDAR scans.
To overcome these limitations, we propose CASSPR as a method to fuse point-based and voxel-based approaches using cross attention transformers. CASSPR leverages a sparse voxel branch for extracting and aggregating information at lower resolution and a point-wise branch for obtaining fine-grained local information.
CASSPR uses queries from one branch to try to match structures in the other branch, ensuring that both extract self-contained descriptors of the point cloud (rather than one branch dominating), but using both to inform the output global descriptor of the point cloud.
Extensive experiments show that CASSPR surpasses the state-of-the-art by a large margin on several datasets (Oxford RobotCar, TUM, USyd).
For instance, it achieves AR@1 of 85.6\% on the TUM dataset, surpassing the strongest prior model by $\sim$15\%. Our code will be publicly available.\footnote{\url{https://github.com/Yan-Xia/CASSPR}}
\end{abstract}

\section{Introduction}

3D place recognition and localization in a city-scale map is a fundamental challenge in allowing autonomous agents to operate effectively in realistic applications, such as autonomous driving \cite{hane20173d, hee2013motion, liu2017robust, liu2021matching, Wu_2021_CVPR, 10124374} and sidewalk or indoor robot navigation \cite{fu2018texture, mur2015orb, xia2021asfm, dou2022coverage, 8463165, 8793716}.

\begin{figure}[ht]
\centering
\includegraphics[width=8cm]{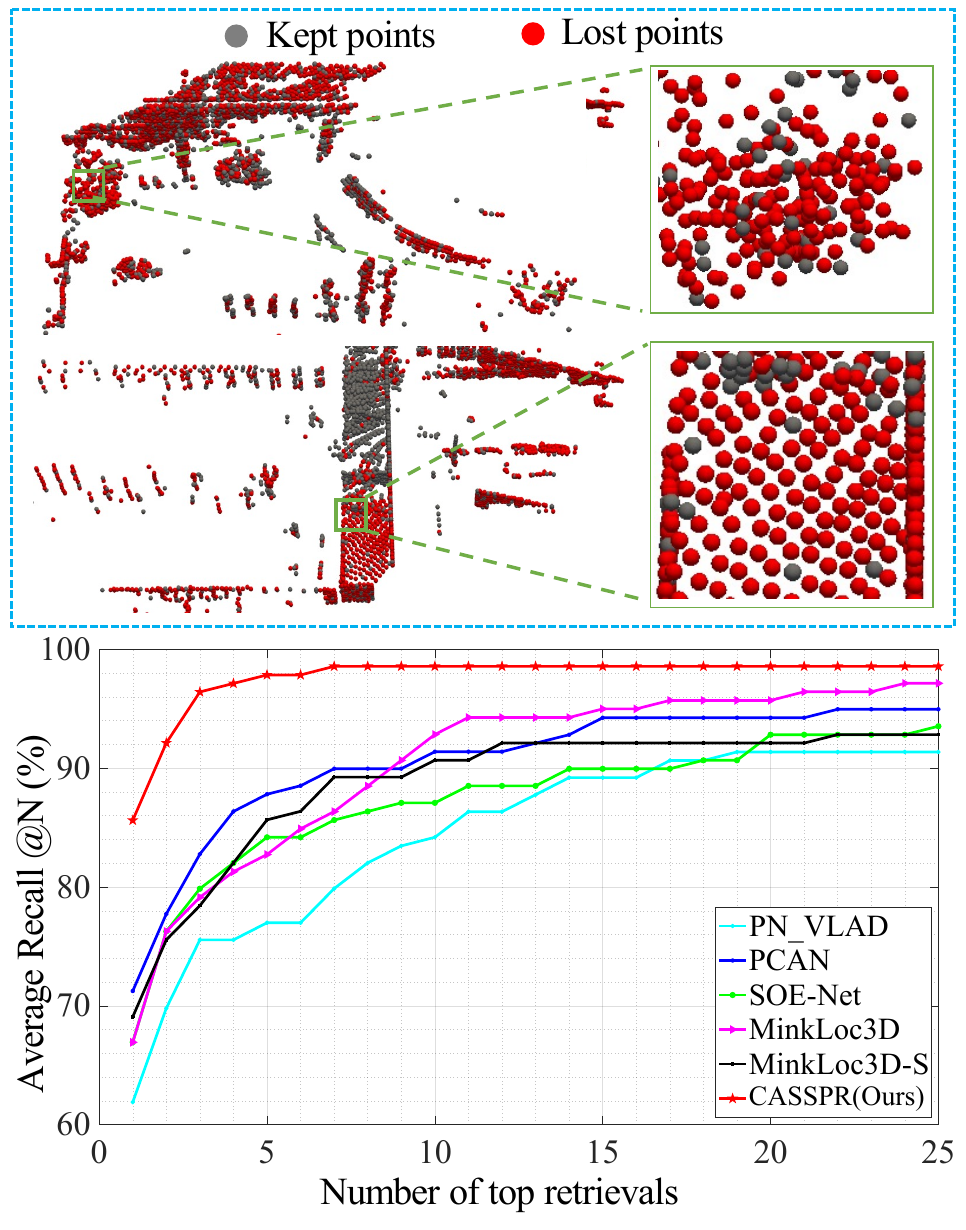}    
\caption{{\bf(Top)} 
Voxelization, which is necessary for sparse 3D convolutions~\cite{zywanowski2021minkloc3dS}, loses a large amount of geometric detail (visualized in red). In this example, 64\% of the points are lost.\\
{\bf(Bottom)} Place recognition performance on the TUM dataset. 
The proposed CASSPR delivers consistently better performance across all top retrieval numbers. Notably, at top 1 retrieval, it outperforms the best baseline by $\sim$15\%. 
}
\label{fig: cover} 
\vspace{-0.5cm}
\end{figure}

While GPS signals can provide reliable geo-location under optimal conditions, they require external satellite information and are bound to fail when there is no direct line-of-sight and the signal is absent, such as in tunnels, parking garages, or among tall buildings and vegetation \cite{arroyo2017topological}.  It is therefore important to develop the capability to perform localization from the on-board sensors only.

Calibrated depth sensors such as LiDAR (producing point clouds) can be used to localize the currently-observed scene by matching it against a pre-built point cloud database. Being focused on the geometry of the scene, point cloud recognition sidesteps several factors that make pure visual (RGB image-based) matching difficult, such as variations in lighting, weather, and season, where the same geometric structure may appear entirely different.

In the past decade, a variety of point-cloud-based solutions has been proposed that achieve excellent performance. They can be classified into two broad categories: point-based descriptors and voxel-based descriptors. 
With several works building on PointNetVlad~\cite{angelina2018pointnetvlad}, point-based descriptors\cite{angelina2018pointnetvlad, zhang2019pcan, liu2019lpd, xia2021soe} are dedicated to improving the global context aggregation (e.g. with pooling or self-attention units).
However, processing points individually can miss some local context, and does not make use of flexible and robust local pattern matchers such as the filters of convolutional neural networks (CNNs).
The voxel-based descriptors~\cite{komorowski2021minkloc3d, zywanowski2021minkloc3dS} generally consist of CNNs leveraging 3D sparse convolutions over a sparse volumetric representation, which can be implemented efficiently using hash tables, most notably with the Minkowski Engine~\cite{choy20194d}.
They achieve good performance with fewer parameters than other models and comparatively fast training.
However, sparse voxels can only represent 1 point per voxel cell. Therefore, for larger cells, several points (possibly containing important geometric information) are necessarily lost -- for example, 64\% of the points in Fig. \ref{fig: cover} (a) are lost during voxelization. Small cells, on the other hand, consume large amounts of memory -- for example, 82.6 GB of GPU memory would be needed to keep 90\% of the points on average~\cite{liu2019point}.
It remains a difficult challenge to capture fine-grained geometric information with point representations, while enjoying the robust pattern matching at multiple context scales of voxel approaches, and without consuming too much memory.

Another challenge is that most previous descriptors have performed place recognition when given dense maps as a reference.
While this is an important setting, accurate point cloud maps may not always be available, in unfamiliar scenes and when an environment changes.
Ideally, one would therefore prefer localization to be performed from sparse single-shot scans, which can be collected during normal online operation of an autonomous system, instead of requiring a carefully-curated densely captured point cloud to be created first.
A single LiDAR such as Ouster OS1-128~\cite{zywanowski2021minkloc3dS} can cover an area between 160 and 200 meters in diameter, collecting a relatively sparse set of points.
Aggregated LiDAR scans are typically at least 10 times denser.  
The larger area and lower density represent a problem for performing place recognition from single LiDAR scans.

To tackle these problems, we propose a novel cross attention transformer for single-scan-based place recognition, named CASSPR, aiming to compensate for the quantization losses and integrating long-range spatial relationships.
The key to maintaining geometric detail even with coarse voxelization is to use a second branch to compute features independently for each point, inspired by PointNet~\cite{qi2017pointnet}.
This branch is free to partition the input space into regions with boundaries that are not necessarily aligned with voxels, and associate different features with each one.
One problem with this strategy is that naively applying a sparse convolution network to such features would still lose geometric detail by removing co-located points due to each voxel only supporting features from a single point.
Therefore, we propose to fuse information from both branches, namely point-wise and sparse voxelized, with a hierarchical cross-attention transformer~\cite{vaswani2017attention} (HCAT).
This transformer
can flexibly aggregate local and global information into sparse voxel features, and do so very efficiently (see Sec.~\ref{sec: time analysis}).
This flexibility and efficiency allows us to surpass the state-of-the-art performance in several challenging point cloud localization tasks (Sec.~\ref{sec: results}).


To summarize, the main contributions of this work are:
\begin{itemize}
    \item We study the  extreme sparsity of single (non-aggregated) LiDAR scans, in the context of voxel and point-based neural networks for place recognition, and analyze the complementary roles of each approach.

    \item We propose a 2-stage hierarchical cross-attention transformer (HCAT) module that is designed to compensate for the shortcomings of point-wise and voxel-wise features, compensating for the loss of the geometric detail caused by the spatial quantization and integrating long-range spatial relationships.
    
    \item We assess the computation and memory trade-offs of the different approaches, finding that our proposal reduces inference time and memory consumption by up to 62\% and 91\%, respectively, when compared to previous attention units used for LiDAR-based place recognition.
    
    \item We conduct extensive experiments on several benchmark datasets, including USyd Campus~\cite{zhou2020developing}, Oxford RobotCar~\cite{maddern20171} and TUM City Campus datasets~\cite{zhu2020tum} and show that the proposed CASSPR greatly improves over the state-of-the-art methods. 
\end{itemize}

\section{Related work}
3D point cloud based place recognition is usually expressed as a 3D retrieval task. Numerous methods were proposed to tackle this task, which can be primarily classified into 2 categories: those based on geometric global descriptors, and those based on plane or object descriptors.

\paragraph{Handcrafted 3D global descriptors.} Most handcrafted global descriptors describe places using global statistics, which have the advantage of not requiring re-training to adapt to different environments and sensor types.
Magnusson \etal~\cite{magnusson2009automatic} split the point cloud into overlapping cells and computed shape properties such as spherical, linear, and several types of planar properties of each cell.
Rohling \etal~\cite{rohling2015fast} propose a fast method of describing places through histograms of point elevation, assuming the sensor had a constant height above the ground plane. 
Scan-Context~\cite{kim2018scan} is a method that exploits a bird-eye view of the point cloud and encodes height information of the surrounding objects for place recognition.
However, handcrafted descriptors cannot improve with more data, placing a hard ceiling on their performance, unlike data-driven methods.

\paragraph{Learning-based 3D global descriptors.} 
With breakthroughs in learning-based image retrieval methods, deep learning of 3D global descriptors for retrieval tasks has become the focus of intense research. 
In PointNetVlad, Uy \etal~\cite{angelina2018pointnetvlad} tackle 3D place recognition with end-to-end learning, using PointNet \cite{qi2017pointnet} to extract local descriptors and then aggregating them globally using NetVlad pooling over 3D points \cite{arandjelovic2016netvlad}. PointNet itself consists of a MLP applied to each point independently, with global max-pooling aggregation of features instead, as well as a predicted linear transform for input points \cite{qi2017pointnet}.
Subsequently, PCAN~\cite{zhang2019pcan} explores an attention mechanism for local features aggregation, discriminating local features that contribute positively. 
LPD-Net~\cite{liu2019lpd} enhances local contextual relationships using graph neural networks, but relying on handcrafted features.
DH3D~\cite{du2020dh3d} introduces a deep hierarchical network to produce more discriminative descriptors, by recognizing places and
refining a 3D pose estimation simultaneously. 
SOE-Net~\cite{xia2021soe} presents a PointOE module introducing orientation encoding into PointNet for generating point-wise local descriptors.
Minkloc3D~\cite{komorowski2021minkloc3d} utilizes a Feature Pyramid Network~\cite{lin2017feature} (FPN) based on 3D sparse tensors with generalized-mean (GeM) pooling~\cite{radenovic2018fine} to compute a compact global descriptor. 
Several methods~\cite{zhou2021ndt, deng2018ppfnet, fan2022svt, xu2023transloc3d, zhang2022rank, ma2022overlaptransformer, barros2022attdlnet, ma2023cvtnet} introduce different transformer networks (i.e. stacked self-attention blocks~\cite{wu2023dropmae, dou2022tore}) for learning long-range contextual features. Compared to OverlapNet~\cite{chen2021overlapnet} and other attention-based methods, including OverlapTransformer~\cite{ma2022overlaptransformer}, AttDLNet~\cite{barros2022attdlnet} and CVTNet~\cite{ma2023cvtnet}, our CASSPR directly applies to raw point clouds, rather than range images projected from LiDAR scans.
The work that is more closely related to ours is Minkloc3D-SI~\cite{zywanowski2021minkloc3dS}, which tackled place recognition with a single 3D LiDAR scan, proposing a non-Cartesian point representation and per-point color intensity information. 
However, 3D voxelization methods inevitably suffer from lost points due to the quantization step (see Fig.~\ref{fig: cover}).

\paragraph{Point-voxel fusion networks.}
Recent works have tried to fuse point-wise networks with voxelized representations. The PV-RCNN~\cite{shi2020pv} architecture in LCDNet~\cite{cattaneo2022lcdnet} adopts a voxel-to-keypoint encoding strategy for 3D object detection, where the features of each keypoint are aggregated by grouping the neighboring voxel-wise features.
In PVCNN~\cite{liu2019point} and PVT~\cite{zhang2022pvt}, the voxel-based features are first transformed back to the domain of the point cloud and then simply added with point-wise features for 3D semantic segmentation. 
Both are different from our method, which does not lose points (either by aggregation or devoxelization), but rather uses them to inform the voxel branch (using cross-attention).

\paragraph{Based on planes or objects.}
Some works have proposed using 3D shapes or object-centric recognition for the task of 3D place recognition. Fernandez \etal \cite{fernandez2013fast} first detect planes in 3D environments and then summarize their pair-wise relationships into a graph. A final geometric consistency test over the planes matches sub-graphs of this structure using an interpretation tree. 
A follow-up work~\cite{fernandez2016scene} utilizes the covariance of the plane parameters instead of the number of points in planes for matching. 
Finman \etal~\cite{finman2015toward} propose an object-based indoor place recognition based on RGB-D cameras.
However, it is only suitable for small, indoor environments with objects of particular classes.

\section{Problem statement} 
\label{sec: problem statement}
We begin by defining the reference map ${ M_\textrm{ref} = {\left \{ m_{i}: i = 1,..., M \right \}}}$ to be a collection of \emph{single-shot} 3D LiDAR scans $m_{i}$. Each scan is tagged with UTM coordinates using GPS/INS at the position where it is captured~\cite{zhou2020developing}.
Let \textit{${Q}$} be a query single LiDAR scan at one timestamp. Then, the single-scan 3D place recognition problem is defined as retrieving the scan \textit{${m^{*}}$} from \textit{${M_\textrm{ref}}$} that is structurally closest to \textit{${Q}$}. 
We aim to design a network $F(\cdot )$ that encodes a single scan as a 3D global descriptor, such that we can retrieve the correct scan ${m^{*}\in M_\textrm{ref}}$ by the nearest-neighbour search in a reference map:
\begin{equation} 
    m^{*}=\underset{m_{i}\in M_\textrm{ref}}{{\rm argmin}}\: d(F(Q), F(m_{i})),
\label{Eq:encoder}
\end{equation}
\noindent 
where $d(\cdot )$ is a distance metric (e.g. the Euclidean distance).
In practice, the global descriptors of all 3D scans are compiled offline in a dictionary. 
Given a new query scan, the nearest neighbor in a dictionary is found efficiently using a KD-tree (e.g. using FAISS~\cite{johnson2019billion}).

\section{Methodology}
\begin{figure*}[ht!]
\centering
\includegraphics[width=\textwidth]{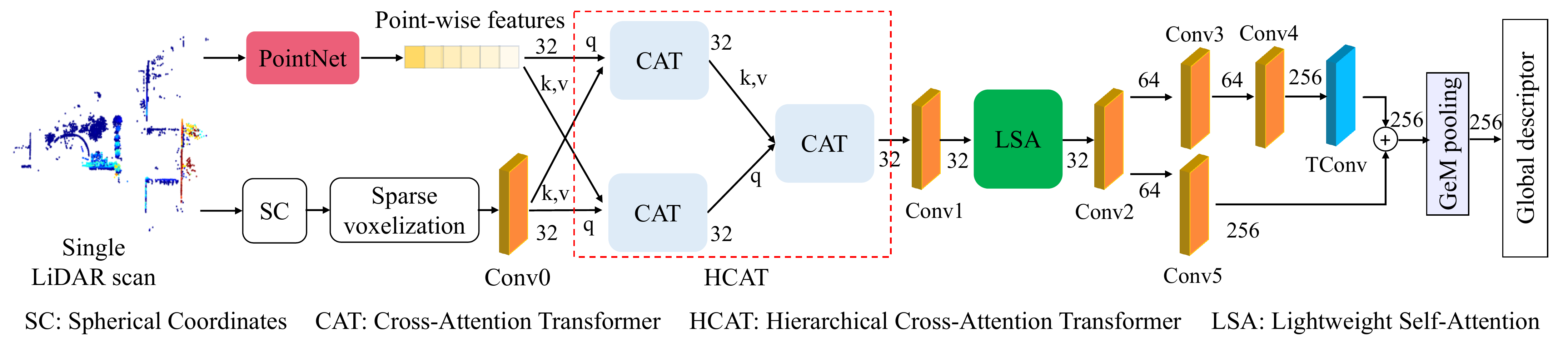}
\caption{The proposed CASSPR architecture. It consists of two parallel branches: one is extracting features from each point independently (top) and another is using sparse convolutions over sparse voxels (bottom). Cross-attention operators use queries from one branch to look up information in the other branch, encouraging the network to focus on the commonalities between these two views. The result is then fused with cross-attention and (light-weight) self-attention, and finally processed with convolutions, deconvolutions (TConv), and GeM pooling following ~\cite{komorowski2021minkloc3d} to generate the final global descriptor. The numerical values (e.g., 32, 64) represent the number of channels of a feature map produced by each block.}
\label{fig:network} 
\end{figure*}

Fig.~\ref{fig:network} shows the network architecture of our CASSPR.
The point branch (top) captures fine-grained geometrical features by using a simplification of a PointNet~\cite{qi2017pointnet}, described in Section~\ref{section:pointnet}.
The voxel branch (bottom) performs convolution on sparse voxels in spherical coordinates (based on Minkloc3D-SI~\cite{zywanowski2021minkloc3dS}) and will be explained in Section~\ref{section: spherical}.
Point-voxel feature fusion is achieved by a Hierarchical Cross-Attention Transformer (HCAT) module (Section~\ref{sec: cross-attention module}).
Finally, lightweight self-attention (LSA) units are used to encode the spatial relationship among local descriptors (Section~\ref{sec:lightweight self-attention unit}), followed by 3D convolutions and pooling.
The training strategy with the loss function is described in Section~\ref{sec:loss function}.

\subsection{Point branch}
\label{section:pointnet}
This branch is a simplification of the PointNet~\cite{qi2017pointnet} without the max-pooling stage.
To make the discussion self-contained, we describe it briefly here.
It consists of a simple 2-layer MLP with 64 channels that operates on each point independently and outputs a 32-dimensional vector for each point.
The input 3D points are transformed via multiplication with an adaptive $3\times 3$ matrix, which allows the PointNet to rotate or stretch the point cloud to align it with a preferred reference frame.
This matrix is predicted by an auxiliary network that also processes each point with a 3-layer MLP, applies max-pooling over the points, and finally applies a 2-layer MLP to output the matrix's 9 elements.

The point branch has the ability to recognize fine-grained geometric features at a global level, without discretization at arbitrary voxels~\cite{qi2017pointnet}.
However, it has limited capacity to flexibly identify patterns in local neighborhoods of points.

\subsection{Sparse spherical voxel branch}
\label{section: spherical}
The density of 3D point clouds in urban scenes collected by LiDAR sensors is uneven, with the area closer to the scanner having a much greater density than far-away regions.
Consequently, 3D voxelization methods based on Cartesian coordinates face a difficult tradeoff: a fine grid captures nearby details but consumes a lot of memory, limiting the spatial range of the grid; a coarse grid allows a greater range but loses detail close to the scanner.
To address this, we use the spherical-voxelization method of MinkLoc3D-S~\cite{zywanowski2021minkloc3dS} to balance the varying density of points in a LiDAR scan.
Each 3D point $(x, y, z)$ is transformed to spherical coordinates $(\gamma,\theta, \varphi)$, with $\gamma=\left\Vert (x,y,z)\right\Vert$ and
\begin{equation} 
\label{Eq:encoder}
\theta=\mathrm{atan}_2(y,x),\quad\varphi=\mathrm{atan}_2(z,\sqrt{x^{2}+y^{2}}).
\end{equation}
\noindent 
The advantage of spherical coordinates is that the volume of each cell grows quadratically with its distance ($\gamma$) from the scanner, taken to be at the origin, resulting in a more even distribution of point density.

Next, we quantize the point cloud with coordinates $(\gamma,\theta,\varphi)$
into a finite number of voxels (Sparse Voxelization block in Fig. \ref{fig:network}).
This process creates a sparse tensor with 3 spatial dimensions, containing the constant 1 in a 3D voxel whenever there is a point there.
This is implemented efficiently using coordinate hash functions with the Minkowski Engine~\cite{choy20194d}.
A 3D convolutional block then aggregates neighboring points, producing a sparse feature map with increasing receptive fields.
We explore how to efficiently fuse the features from different branches in the next sections. 
Note that, while the voxelization allows for robust 3D sparse convolutions to extract geometric information from a wide receptive field, it necessarily loses many details from the point cloud (visualized in Fig.~\ref{fig: cover}, top).

\subsection{Hierarchical cross-attention transformer}
\label{sec: cross-attention module}
\begin{figure}[t]
\centering
\includegraphics[width=1.0\columnwidth]{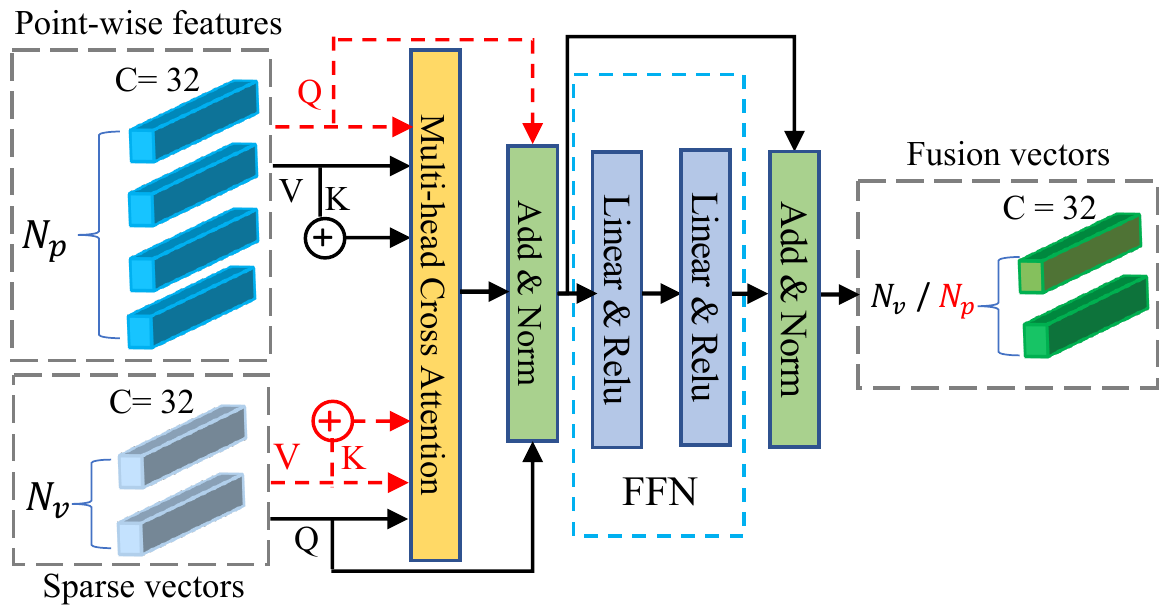}  
\caption{The architecture of the Cross-Attention Transformer.
See Sec. \ref{sec: cross-attention module} for details.
} 
\label{fig: HCAT} 
\end{figure}

To efficiently exploit the relationship between the volumetric branch and the point branch, we further propose a Hierarchical Cross-Attention Transformer (HCAT) to fuse the features from the two branches in a unified model. As shown in Fig.~\ref{fig:network}, the HCAT consists of three Cross Attention Transformers (CAT), each having a different role. The first CAT takes the sparse-voxel features as the Query and the point-wise features as the Key and Value. It extracts point-wise features with reference to the sparse feature maps and outputs sparse tensors that are informed by the point features. 
Conversely, another CAT produces improved point-wise features by taking the point features as the Query and the sparse-voxel features as the Key and Value.
Finally, the aim of the third CAT is to fuse both enhanced sparse tensors and enhanced point-wise features. Fig.~\ref{fig: HCAT} shows the architecture of the HCAT in detail.
The advantage of this scheme, as opposed to simply using a single CAT with concatenated inputs from both branches, is to ensure that each branch is useful in isolation -- as it must be used as the sole source of Queries for a single CAT calculation, but also independently as the sole source of Keys and Values -- thus preventing one branch from dominating over the other.

Each CAT is a residual module, consisting of two sub-layers: Multi-Head Cross-Attention (MHCA) and Feed-Forward Network (FFN). The feed-forward network includes two linear transformation layers with the ReLU activation function. Formally, each CAT can be formulated as:

\begin{equation}
\begin{aligned}
\mathbf{F}_{CA} & = \operatorname{CAT}(\mathbf{Q}, \mathbf{K}, \mathbf{V}) \\
& =\widetilde{\mathbf{F}}_{CA}+\operatorname{FFN}\left(\widetilde{\mathbf{F}}_{CA}\right), \\
\widetilde{\mathbf{F}}_{CA}& =\mathbf{Q}+\operatorname{MHCA}\left(\mathbf{Q}, \mathbf{K}, \mathbf{V}\right),
\end{aligned}
\end{equation}
\noindent
where $\mathbf{Q} = F_{s} \in \mathbb{R}^{N_{s} \times d}$ is the query, $\mathbf{K} = \mathbf{V} = F_{p} \in \mathbb{R}^{N_{p} \times d}$  are the key and value. 

In the MHCA layer, cross-attention is performed by projecting the $\mathbf{Q}$, $\mathbf{K}$, and $\mathbf{V}$ with $h$ heads (we use $h=2$).
Specially, we first calculate the weight matrix with scaled dot-product attention \cite{vaswani2017attention}  for the Query, Key, and Value:
\begin{equation}
\operatorname{Attention}\left(\mathbf{Q}, \mathbf{K}, \mathbf{V}\right)=\operatorname{S}\left(\frac{\mathbf{Q}+ \mathbf{K}^{T}}{\sqrt{d_{k}}}\right) \mathbf{V},
\end{equation}
where $\operatorname{S}$ is a Softmax function outputting attention weights.
Next, we calculate the values for $h$ heads and concatenate them together:
\begin{align}
\text {Multi-Head}&(\mathbf{Q,K,V})=\left[ \text{ head}_{1}, \ldots, \text {head}_{h}\right] \mathbf{W}^{O},\\
\text {head}_{i}&=\text {Attention}\left(\mathbf{Q} \mathbf{W}_{i}^{Q}, \mathbf{K} \mathbf{W}_{i}^{K}, \mathbf{V} \mathbf{W}_{i}^{V}\right),
\end{align}
\noindent
where $\mathbf{W}_{i}^{\{Q,K,V,O\}}$ are learnable parameters.

With the above HCAT, point-wise features from the point branch can attend to the sparse tensors from the voxel-based branch, and vice-versa, thus compensating for the loss of points due to the voxel quantization.

\subsection{Lightweight self-attention unit}
\label{sec:lightweight self-attention unit}
Before going into details concerning the Lightweight Self-Attention (LSA) unit, we give a short review of the standard self-attention used in previous 3D place recognition or point Transformers \cite{zhao2021point, guo2021pct}.
Given a set of points with features $D_{L}=\left \{ (p_{i},f_{i}) \right \}_{i=1}^{N}$, where $p_{i}$ is the coordinate of the $i$-th point (of a total of $N$) and $f_{i} \in \mathbb{R}^{C}$ is the feature of $p_{i}$.
A dot-product self-attention (DPSA) unit on $p_{i}$ can be formulated as follows:
\begin{equation}
\small
\begin{split}
    \operatorname{DPSA}\left({p_{i}} \right) & = \sum_{j \in \chi(i)} \operatorname{S}[\delta(p_{i}-p_{j}), f_{i}]
    \varphi(f_{j}),
    \label{Eq:attention_map}
\end{split}
\end{equation}
\noindent
where $\delta$ is a linear positional encoding function, $\varphi$ is a learnable value projection layer, and $\chi(i)$ denotes the indices of neighboring points of $p_{i}$.
  
However, computing $\operatorname{DPSA}$ directly by multiplying two $N \times C $ matrices is expensive, with $O(N^{2})$ space complexity and $O(N^{2.34})$ time complexity \cite{alman2021refined}. 
Additionally, nearest neighbors search with a KD-tree has $O(N\log N)$ space complexity~\cite{park2022fast}.
In this work, we reduce the training and inference time$/$memory consumption using a Lightweight Self-Attention (LSA) unit, inspired by~\cite{park2022fast}, which we describe next.

{\bf Efficient positional encoding.} We first voxelize the input point cloud to a set of $M$ triplets $\mathbb{V}= \left \{ (v_{i},g_{i}, c_{i}) \right \}_{i=1}^{M}$, including the $i$-th voxel coordinate $v_{i}$, the corresponding voxel feature $g_{i}$, and the centroid coordinate $c_{i}$ of this voxel. Note that $M< N$ since the sparse voxelization step will lose most points. 
LSA on $c_{i}$ is formulated from Eq.~\ref{Eq:attention_map}:
\begin{equation}
    \operatorname{LSA}\left(c_{i} \right) = \sum_{j \in \tau(i)} \operatorname{S}\left [\delta(c_{i} - c_{j}), g_{i} \right ]\varphi(g_{j}),
    \label{Eq:light_attention_map}
\end{equation}
\noindent
where $\tau(i)$ represents the neighbor voxel indexes of $c_{i}$.

\begin{figure}[t]
\centering
\includegraphics[width=1.0\columnwidth]{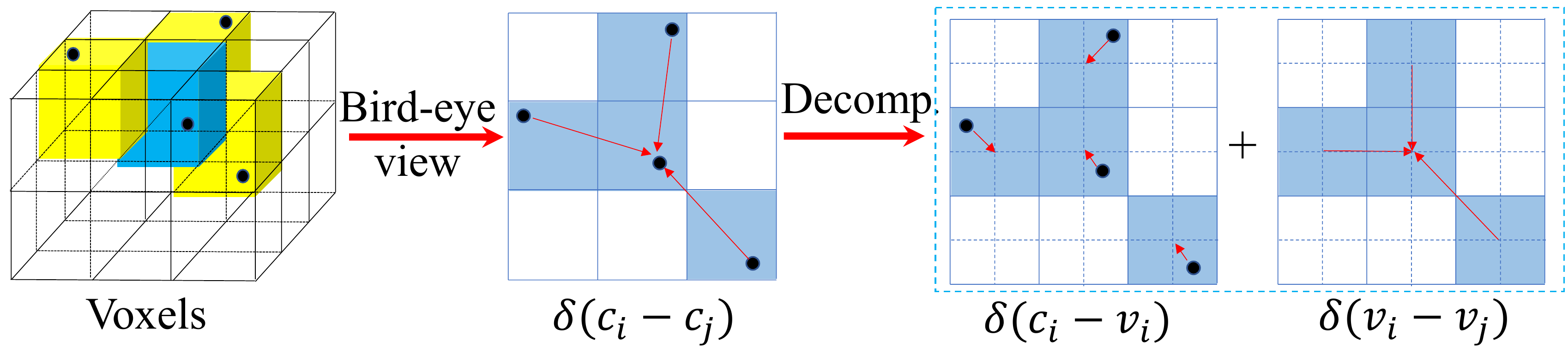}    
\caption{Decomposition of positional encoding. We decompose the positional encoding $\delta(c_{i}-c_{j})$ to $\delta(c_{i}-v_{i})$ and $\delta(v_{i}-v_{j})$. This decomposition can reduce the space complexity from $ O(MKD)$ to $ O(MD + KD)$.} 
\label{fig: positional} 
\end{figure}

Finding neighboring $K$ voxels via voxel hashing will be quick since it only has $O(M)$ time complexity. However, implementing  $\delta(c_{i},c_{j})$ directly as a linear function requires $O(MKD)$ space complexity. To alleviate this problem, a coordinate decomposition approach is used inspired by ~\cite{park2022fast}. Given a query voxel $(v_{i},g_{i}, c_{i})$ and a nearest neighbor voxel $(v_{j},g_{j}, c_{j})$, the relative position encoding between $c_{i}$ and $c_{j}$ can be decomposed as follows:
\begin{equation}
    \delta(c_{i}-c_{j}) = \delta \left [(c_{i} - v_{i}) + (v_{i} - v_{j}) - (c_{j} - v_{j}) \right].
    \label{Eq:decomposition}
\end{equation}
\noindent
As seen in Eq.~\ref{Eq:decomposition}, the memory-consuming $\delta(c_{i}-c_{j})$ is decomposed in three parts,
since $\delta$ is a linear operator.
The space complexity of $\delta(c_{i},c_{j})$ is reduced from $ O(MKD)$ to $ O(MD + KD)$ because the space complexity of $\delta(c_{i}-v_{j})$ and $\delta(v_{i}-v_{j})$ will be $O(MD)$ and $O(KD)$.
The decomposition is illustrated in detail in Fig.~\ref{fig: positional}.

Unfortunately, using the softmax function in Eq.~\ref{Eq:light_attention_map} will often result in degenerate weights due to the sparsity of the inputs.
The input size is variable and often an input only contains a single point or none.
When $|\tau_{i}|$ is 1 (i.e. a single point in the neighborhood), the attention weights are normalized into the constant 1, so LSA will be downgraded to a linear function $\varphi$.
To prevent this, we replace the softmax with cosine similarity in Eq.~\ref{Eq:light_attention_map}, rewriting it as:
\begin{equation}
    \operatorname{LSA}\left(c_{i}\right)=\sum_{j\in\tau_{i}}\frac{\delta^{T}(c_{i}-c_{j})g_{i}}{\left\Vert \delta(c_{i}-c_{j})\right\Vert \left\Vert g_{i}\right\Vert }\varphi(g_{j})
    \label{Eq:attention_map_new}
\end{equation}
After the LSA operation, the result is decoded with 3D convolutional and deconvolution operations, with a coarse residual block inspired by Feature Pyramid Networks~\cite{lin2017feature} (illustrated in Fig. \ref{fig:network}).
Finally, generalized mean pooling (GeM) \cite{radenovic2018fine} aggregates over all voxels to obtain a single 256-dimensional embedding vector. GeM is defined as $e_{k}=\left(\frac{1}{n}\sum_{j}^n h_{kj}^{p}\right)^{1/p}$, given the $k$-th feature of the $j$-th voxel $h_{kj}$ and a learnable exponent parameter $p$.

\section{Loss function}
\label{sec:loss function}
Similarly to Zywanowski et al. \cite{zywanowski2021minkloc3dS}, we use a triplet margin loss based on the batch hard negative mining approach. 
A batch of scans with size $n$ is built by sampling $n/2$ structurally similar scans.
The batch is then used as the input of our network to compute global descriptors.
Notably, two $n \times n$ boolean masks are computed, one indicating structurally similar pairs and the other structurally dissimilar pairs. Informative training triplets are constructed based on the boolean masks. Additionally, we only mine the hardest positive and 
the hardest negative scans in one batch.  The hard mining triplet loss can be formulated as:
\begin{equation}
L= \operatorname{max}\left(m + d\left(\psi_{a}, \psi_{p}\right)-d\left(\psi_{a}, \psi_{n}\right), 0\right)
\end{equation}
\noindent
where $\psi_{a}$ is an anchor point cloud, $\psi_{p}$ a hard positive point cloud (structurally similar to the anchor), $\psi_{n}$ a hard negative point cloud (structurally dissimilar to the anchor). $m$ is a pre-defined margin parameter.

\section{Experiments}
\subsection{Datasets}
To showcase the performance of our CASSPR on single LiDAR scan data, we first train and evaluate CASSPR on two public LiDAR datasets: USyd Campus~\cite{zhou2020developing} and TUM City Campus~\cite{zhu2020tum} dataset. To test the generalization ability and compare with other methods fairly, we also train and evaluate on Oxford RobotCar~\cite{maddern20171}. 
Different with USyd and TUM dataset, the point clouds in Oxford RobotCar are generated from 2D scans accumulated over time.

{\bf USyd Campus Dataset.}
The USyd Campus Dataset (USyd)~\cite{zhou2020developing} contains data collected by a small electric car driving the same route around the university of Sydney campus over 50 weeks in varying weather conditions. The array of sensors includes a Velodyne VLP-16 LiDAR, six cameras, and GPS/IMU. 
Following~\cite{zywanowski2021minkloc3dS}, we split consecutive LiDAR scans at intervals of 5 meters traveled distance, resulting in about 735 scans each run. Each scan covers 100 meters in diameter.
The point clouds are downsampled using the voxel grid filter to 4096 points. Notably, the ground plane is not removed.

{\bf Oxford RobotCar Dataset.}
The Oxford RobotCar dataset~\cite{maddern20171} consists of data collected by a SICK LMS-151 2D LiDAR scanner recorded over a year, with a total length over \SI{1000}{km}.
Point clouds in Oxford RobotCar are generated from 2D scans accumulated over time to create a unified 3D map.
The 3D map is divided into submaps with \SI{20}{m} length, and each submap includes exactly 4096 equally distributed points. 

{ \bf TUM City Campus Dataset.}
The TUM City Campus (TUM) dataset ~\cite{zhu2020tum} contains two recordings taken at the city campus of the Technical University of Munich.
Two Velodyne HDL-64E LiDAR sensors are mounted on Mobile Distributed Situation Awareness (MODISS).
The acquisition resulted in more than 10,500 scans for each run.
A single scan includes \SI{130}{K} points per rotation and covers a large area of 120 meters in diameter. 
The point cloud scans are split roughly every 5 meters without overlapping based on the scanner positions. 
Ground removal is applied to remove uninformative patterns. We then downsample the point clouds to 4096 points using the voxel grid filter. 

{ \bf In-house datasets.}
We validate the generalization capabilities of networks trained only on Oxford RobotCar by utilizing three additional in-house datasets collected in a university sector (U.S.), a residential area (R.A.), and a business district (B.D.) with a sensor Velodyne-64 LiDAR~\cite{angelina2018pointnetvlad}. The datasets contain single-scan sequences of lengths \SI{10}{km}, \SI{8}{km}, and \SI{5}{km}, respectively.

\subsection{Evaluation criteria} 
Following~\cite{angelina2018pointnetvlad}, Average Recall at Top $N$ (AR@N) is used as an evaluation metric, which means the location is correctly recognized if the $N$ most similar scans matched from the database contain at least one location within the distance $d$ from the query. Top $1$ counts the number of times the first match from the database matches the query location.
We also present the Average Recall for Top $1\%$ (AR@1\%) results for comparison to the state-of-the-art solutions.
For the TUM dataset, we regard the retrieved scan as a correct match if the distance is within $d = 5m$.
For the USyd, Oxford RobotCar, and In-house datasets, the distance is set as 10m, 25m, and 25m, respectively.

\subsection{Results}
\label{sec: results}
\subsubsection{Comparisons on the USyd and TUM Dataset.}

\begin{table}[t!]
\centering
\caption{Average recall ($\%$) at top 1$\%$ (@1$\%$) and top 1 (@1) for each model trained on the USyd and TUM dataset. * indicates the results reported in MinkLoc3D-S~\cite{zywanowski2021minkloc3dS}.}
\resizebox{0.46\textwidth}{!}{
\begin{tabular}{ccccc} 
\hline
\multicolumn{1}{l}{\multirow{2}{*}{}} & \multicolumn{2}{c}{USyd dataset} & \multicolumn{2}{c}{TUM dataset}  \\
\multicolumn{1}{l}{}                  & ~ ~AR @1\%    & AR @1            & AR @1\%       & AR @1            \\ 
\hline
PointNetVlad
& 81.7          & 60.7             & 76.3          & 61.9             \\
PCAN
& 86.4          & 68.7             & 87.8          & 71.2             \\
SOE-Net
& 78.9          & 52.8             & 83.5          & 66.9             \\
MinkLoc3D
& 98.1*          & 91.7*             & 82.7          & 66.9             \\
MinkLoc3D-S
 & 98.8*          & 93.9*             & 85.7          & 69.1             \\
CASSPR 
& \textbf{98.9} & \textbf{97.6}    & \textbf{97.1} & \textbf{85.6}    \\
\hline
\end{tabular}}
\label{table:usyd_experiment}
\end{table}

We compare our CASSPR with the state-of-the-art methods: PointNetVlad~\cite{angelina2018pointnetvlad}, PCAN~\cite{zhang2019pcan}, SOE-Net~\cite{xia2021soe}, MinkLoc3D~\cite{komorowski2021minkloc3d}, and MinkLoc3D-S~\cite{zywanowski2021minkloc3dS}.
For a fair comparison, we re-trained and tested on the USyd dataset using publicly-available code.
Table~\ref{table:usyd_experiment} (left)  shows the top 1\% and top 1 recall of each method on the USyd datasets. 
CASSPR achieves the best performance of 98.9\% / 97.6\% at AR@1\% / AR@1, exceeding the performance of the previous SOTA MinkLoc3D-S by 3.9\% on AR@1. 
It demonstrates that CASSPR can generate more discriminative global descriptors compared with point-based or voxel-based baselines.

We also conduct experiments on the TUM dataset.
The results in Table~\ref{table:usyd_experiment} (right) show that the proposed CASSPR outperforms others significantly.
Notably, CASSPR achieves the recall of 97.1$\%$ at top 1$\%$, exceeding the recall of the current state-of-the-art method by 9.3$\%$. 
Furthermore, our CASSPR achieves an average recall of 85.6$\%$ at top 1, which has a significant advantage ($16.5\%$ ) over MinkLoc3D-S.
Fig.~\ref{fig: cover}~(\textbf{Bottom}) shows the recall curves of PointNetVLAD, PCAN, SOE-Net, Minkloc3D, MinkLoc3D-S, and ours for the top 25 retrieval results. The plot demonstrates a consistently superior performance of CASSPR over the compared baselines. 
Comparing the results on the USyd and TUM benchmarks, we notice CASSPR performs much better on the latter. This might be caused by the fact that the TUM dataset has less data compared to the USyd dataset with only two recorded runs. 

\subsubsection{Comparisons on the Oxford RobotCar datasets.}
To further demonstrate the capabilities of our CASSPR, we also conduct experiments on benchmark datasets introduced in \cite{angelina2018pointnetvlad}.
Following the baseline networks proposed in \cite{angelina2018pointnetvlad, zhang2019pcan, liu2019lpd, hui2022efficient, xia2021soe, komorowski2021minkloc3d, zhou2021ndt, hui2021pyramid, fan2022svt, hou2022hitpr}, we train and evaluate on the Oxford RobotCar. Three in-house datasets are used to verify the generalization ability of models on unseen scenarios. 
We use the same settings in Minkloc3D, except that we add the LSA unit followed by each 3D convolution layer. The HCAT is removed because the spherical representation is not suitable for 2D aggregated scans, as verified in \cite{zywanowski2021minkloc3dS}.

We compare CASSPR with the state-of-the-art methods, including PointNetVLAD~\cite{angelina2018pointnetvlad}, PCAN~\cite{zhang2019pcan}, LPD-Net~\cite{liu2019lpd}, EPC-Net~\cite{hui2022efficient}, SOE-Net~\cite{xia2021soe}, Minkloc3D~\cite{komorowski2021minkloc3d}, NDT-Transformer~\cite{zhou2021ndt}, PPT-Net~\cite{hui2021pyramid}, SVT-Net~\cite{fan2022svt}, and HiTPR~\cite{hou2022hitpr}.
The dimensions of all global descriptors are set to 256. 
The evaluation results are shown in Table.~\ref{table:oxford}.
CASSPR achieves state-of-the-art results on the Oxford RobotCar, with 2.6\% improvements at AR@1 over MinkLoc3D.  
Compared with NDT-Transformer, PPT-Net, SVT-Net, and HiTPR, which are all based on Transformers, our CASSPR still achieves a remarkable improvement. Compared with PCAN and SOE-Net, CASSPR exceeds the recall of PCAN and SOE-Net by 26.5\% and 6.2\% at AR@1, respectively. This suggests that the proposed hierarchical attention with a point branch is more effective than the attention strategies used in PCAN and SOE-Net.

In addition, CASSPR surpasses other methods significantly on the in-house datasets, despite the existence of a large domain shift.
This demonstrates that the global descriptors generated by CASSPR can generalize better than the previous state-of-the-art methods.

\begin{table}[t]
\centering\footnotesize
\caption{Average recall ($\%$) at top 1$\%$ (@1$\%$) and top 1 (@1) for each of the models trained on the Oxford RobotCar. Our CASSPR achieves the best performance on all benchmarks.}

\setlength\tabcolsep{3.5pt}  

\begin{tabular}{lcccccccc} 
\toprule
 & \multicolumn{2}{c}{Oxford~}                                     & \multicolumn{2}{c}{U.S.}                                        & \multicolumn{2}{c}{R.A.}                                        & \multicolumn{2}{c}{B.D.}                                         \\
AR @ \{1, 1\%\}  & 1                         & 1\%                        & 1                         & 1\%                        & 1                          & 1\%                        & 1                          & 1\%                         \\ 
\midrule
PN-VLAD~\cite{angelina2018pointnetvlad}                          & 62.8                           & 80.3                           & 63.2                           & 72.6                           & 56.1                           & 60.3                           & 57.2                           & 65.3                            \\
PCAN~\cite{zhang2019pcan}                                  & 69.1                           & 83.8                           & 62.4                           & 79.1                           & 56.9                           & 71.2                           & 58.1                           & 66.8                            \\
LPD-Net~\cite{liu2019lpd}                               & 86.3                           & 94.9                           & 87.0                           & 96.0                           & 83.1                           & 90.5                           & 82.5                           & 89.1                            \\
EPC-Net~\cite{hui2022efficient}                               & 86.2                           & 94.7                           & -                              & 96.5         & \textbf{-}                     & 88.6                           & -                              & 84.9                            \\
HiTPR~\cite{hou2022hitpr} & 86.6 & 93.7 & 80.9 & 90.2 & 78.2 & 87.2 & 74.3 & 79.8 \\
SOE-Net~\cite{xia2021soe}                               & 89.4                   & 96.4                   & 82.5                   & 93.2                   & 82.9                   & 91.5                   & 83.3                   & 88.5                             \\
MinkLoc3D~\cite{komorowski2021minkloc3d}                             & 93.0                           & 97.9                           & 86.7                           & 95.0                           & 80.4                           & 91.2                           & 81.5                           & 88.5                            \\
NDT-T~\cite{zhou2021ndt}                       & 93.8                           & 97.7                           & -                              & -                              & -                              & -                              & -                              & -                               \\
PPT-Net~\cite{hui2021pyramid}                               & 93.5                           & \textcolor{black}{98.1}                           & \textcolor{black}{90.1} & \textcolor{black}{97.5} & 84.1                           & \textcolor{black}{93.3}         & 84.6        & 90.0          \\
SVT-Net~\cite{fan2022svt}                               & 93.7                           & 97.8                           & \textcolor{black}{90.1} & 96.5         & \textcolor{black}{84.3}         & 92.7                           & \textcolor{black}{85.5} & \textcolor{black}{90.7}  \\
MinkLoc3D-S~\cite{zywanowski2021minkloc3dS}         & 92.8                   & 81.7                   & 83.1                      & 67.7                      & 72.6                      & 57.1                      & 70.4                      & 62.2                       \\
\midrule
CASSPR (Ours)                                 & \textbf{95.6}         & \textbf{98.5}         & \textbf{92.9}         & \textbf{97.9}                           & {\textbf{89.5}} &  \textbf{94.8} & \textbf{87.9}                           & \textbf{92.1}                            \\
\bottomrule
\end{tabular}
\label{table:oxford}
\end{table}

\subsection{Ablation study}
To assess the relative contribution of each module,
we remove the LSA and the HCAT (including the point branch) from our network one by one, denoted as CASSPR\_HCAT and CASSPR\_LSA, respectively.
We also switch the key/value and query for the fusion unit of HCAT, namely the point branch acting as a query. and the voxel branch as a key and value, denoted as CASSPR\_Switch.
All networks are trained on the TUM dataset, with results shown in Table \ref{tab:ablation study}.
CASSPR\_HCAT outperforms MinkLoc3D-S in recall by 8.6\%,
indicating the proposed HCAT
is a crucial part of a successful fusion strategy. 
The LSA unit brings significant improvements on the average recall metric (7.9\%), when compared against MinkLoc3D-S performance.
The results are consistent with our expectation that compensating for quantization losses and introducing long-range contexts is essential for generating a more discriminative global descriptor. 
Combining both modules achieves the best performance, improving the performance by 11.4\% and 16.5\%. 
We also observe inferior performance of CASSPR\_Switch to the proposed method when changing the sequence of key/value and query in the second stage. Specifically, CASSPR\_Switch achieves top 1\% recall of 98.1\% on the USyd dataset and 89.2\% on the TUM dataset, lower than the accuracy achieved by the proposed CASSPR of 98.7\% and 97.1\% respectively. We thus conclude the voxel branch plays a more significant role compared to the point branch when fusing the features.

\begin{figure}
\centering
\includegraphics[width=0.47\textwidth]{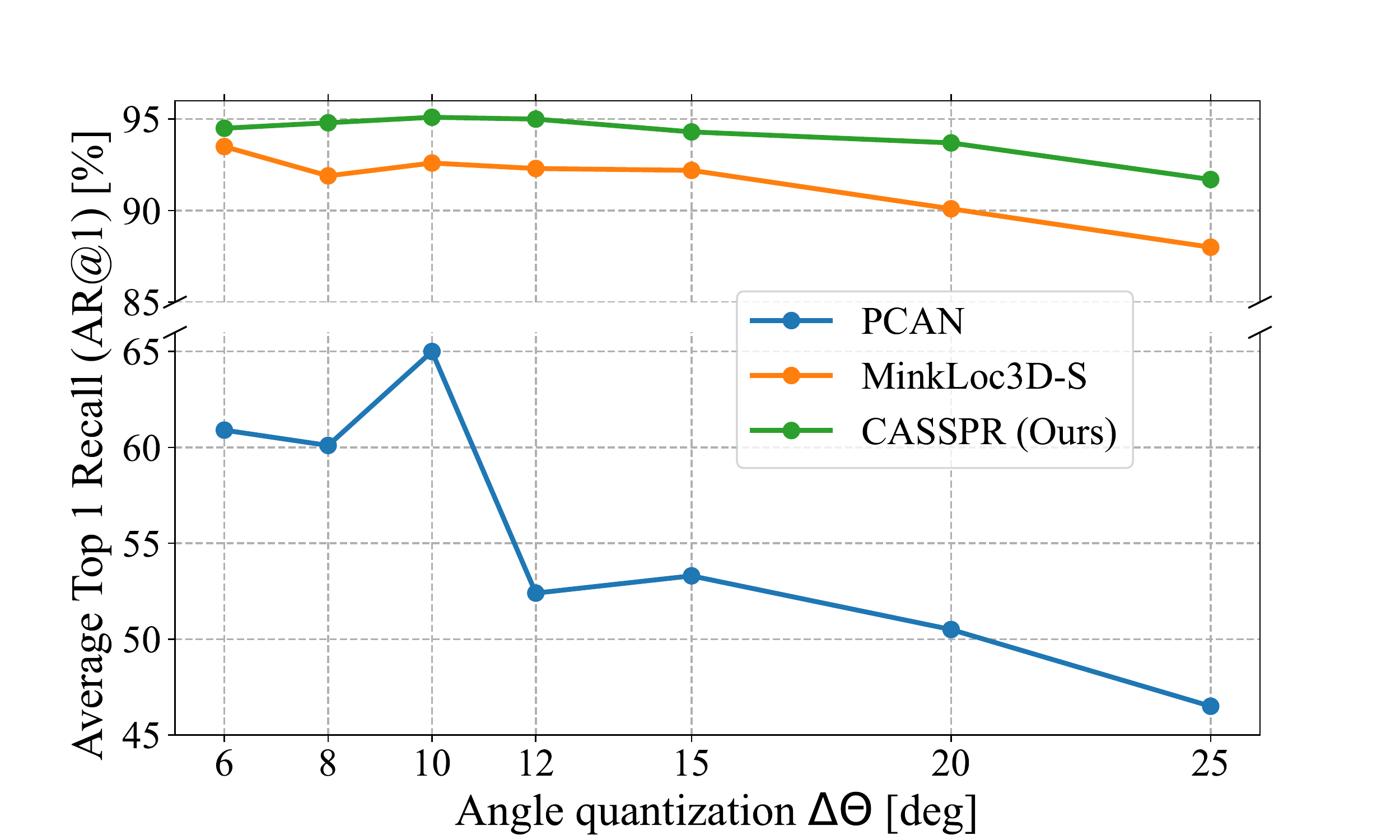} 
\caption[The impact of increasing azimuthal angle quantization $\Delta \theta$ on the average top 1 recall (AR@1) for MinkLoc3D-S~\cite{zywanowski2021minkloc3dS}, PCAN~\cite{zhang2019pcan} and our CASSPR on the USyd sequences~\cite{zhou2020developing}.]{The impact of increasing azimuthal angle quantization $\Delta \theta$ on the average top 1 recall (AR@1) for MinkLoc3D-S~\cite{zywanowski2021minkloc3dS}, PCAN~\cite{zhang2019pcan} and our CASSPR on the USyd sequences~\cite{zhou2020developing}. While point-based PCAN is significantly affected by larger voxelization ranges, CASSPR remains robust across a wide quantization spectrum. Our method is also consistently more accurate than MinkLoc3D-S.} 
\label{fig:quant_radial}
\end{figure}

\paragraph{Impact of sparsity.}
We also studied how performance changes as we vary the sparsity of the point clouds. 
The horizontal scanning angle $\theta$ (presented in Sec.~\ref{section: spherical}) is chosen to determine the number of points fed into the network. Specifically, more points would be lost as $\Delta \theta$ increases. As we vary $\Delta \theta$ the other two parameters are kept constant at their default values, $\Delta r = 2.5$ and $\Delta \varphi = 1.875$.
Fig.~\ref{fig:quant_radial} shows the influence of increasing the quantization parameter $\Delta \theta$ 
on the performance of our CASSPR and two baselines from both point-based and voxel-based communities, PCAN~\cite{zhang2019pcan} and MinkLoc3D-S~\cite{zywanowski2021minkloc3dS}. We randomly choose 10 sequences from the USyd dataset~\cite{zhou2020developing} for the study. We split all scans into train and test sets analogously to~\cite{zywanowski2021minkloc3dS}. While MinkLoc3D-S naturally quantizes input point clouds in order to construct sparse input tensors, PCAN relies entirely on the input scan data. Therefore, to fairly compare point-based and voxel-based approaches, we manually quantize point clouds based on the chosen quantization parameters before feeding them to PCAN.
PCAN is significantly affected by the point cloud quantization with a larger azimuthal angle (Fig.~\ref{fig:quant_radial}). In comparison to the voxel-based MinkLoc3D-S, our CASSPR shows consistently higher accuracy and robustness to increasingly sparse data. Specifically, a larger horizontal angular size (6 to 25 degrees) results in the AR@1 decrease of $\approx$ 3\%, while the impact is more significant for MinkLoc3D-S with $\approx$ 6\%. We attribute this effect to the interplay between the point-based and voxel-based branches of our method. While the voxel-based branch becomes weaker and less informative with increasing sparseness, the point-based branch plays a more influential role thanks to the mediation of the proposed hierarchical cross-attention unit.

\begin{table}
\centering
\caption{Ablations of the hierarchical cross-attention transformer (HCAT) and lightweight self-attention (LSA) on the TUM dataset.}
\resizebox{0.47\textwidth}{!}{
\begin{tabular}{cccccc} 
\toprule
Method     & HCAT & LSA & Ave recall @1\% & Ave recall @1  \\ 
\midrule
MinkLoc3D~\cite{komorowski2021minkloc3d} &   &   & 82.7    & 66.9 \\ 
MinkLoc3D-S~\cite{zywanowski2021minkloc3dS}          &          &          & 85.7    & 69.1       \\
CASSPR\_HCAT    & \checkmark         &          & 94.3       & 77.7      \\
CASSPR\_LSA    &          & \checkmark         & 93.6   & 84.9      \\
CASSPR\_Switch & \checkmark         & \checkmark         & 89.2    & 77.7 \\
CASSPR (Ours) & \checkmark         & \checkmark         & \textbf{97.1}    & \textbf{85.6}       \\
\bottomrule
\end{tabular}}
\label{tab:ablation study}
\end{table}

\subsection{Computational cost analysis}
\label{sec: time analysis}

\begin{table}[t]
\centering
\caption{Computational cost requirements of different 3D global descriptors on the TUM dataset. * indicates the consumption including the HCAT and an LSA.}
\label{tab: time results}
\resizebox{0.47\textwidth}{!}{
\begin{tabular}{lcccccc} 
\toprule
\multicolumn{1}{c}{\multirow{2}{*}{Methods}} & \multicolumn{2}{c}{Parameters (M) }        & \multicolumn{3}{c}{Time Usage (ms)} & Performance    \\ 
\multicolumn{1}{c}{}                         & \multicolumn{1}{l}{Total} & Attention      & Total        & Backbone & Attention & AR @1\%        \\ 
\midrule
PointNetVLAD~\cite{angelina2018pointnetvlad}                                & 19.8                      & N/A              & 12.6         & 12.6     & N/A         & 76.3           \\
PCAN~\cite{zhang2019pcan}                                        & 20.4                      & 0.6            & 66.8         & 27.8         & 39.0           & 87.8           \\
SOE-Net~\cite{xia2021soe}                                     & 19.4                      & 1.6            & 66.9         &   29.8       &    37.1       & 83.5           \\
MinkLoc3D-S~\cite{zywanowski2021minkloc3dS}                                 & \textbf{1.1}              & N/A              & \textbf{6.0} & 6.0      & N/A         & 85.7           \\
CASSPR (Ours)                                 & 3.8                       & \textbf{0.057*} & 29.7         & 15.6      & 14.1*     & \textbf{97.1}  \\
\bottomrule
\end{tabular}
}
\end{table}

In this section, we analyze the required computational resources of different global descriptors in terms of the number of parameters and time efficiency. For a fair comparison, all methods are tested on the TUM dataset with a single NVIDIA V100 (32G) GPU.
As shown in Table \ref{tab: time results}, our CASSPR takes \SI{29.7}{ms} to encode one scan into a global descriptor with only \SI{3.8}{M} parameters. 
Although CASSPR is a point-voxel fusion architecture, it has lower trainable parameters compared with the pure point-based methods, including PointNetVLAD, PCAN, and SOE-Net. 
For inference time, CASSPR is faster than the attention-based methods PCAN and SOE-Net per scan (\SI{29.7}{ms} vs. \SI{66.8}{ms}).
In addition, our attention units, including the HCAT and LSA, are significantly more lightweight (91\% improvement in memory) and faster (62\% improvement in inference time).
However, compared with pure voxel-based MinkLoc3D-S, CASSPR has more parameters and running time due to the extra point branch (\SI{5}{ms}), HCAT module (\SI{12.7}{ms}) and LSA (\SI{1.4}{ms}) unit. 

\begin{figure}[tbp]
\centering
\includegraphics[width=0.47\textwidth]{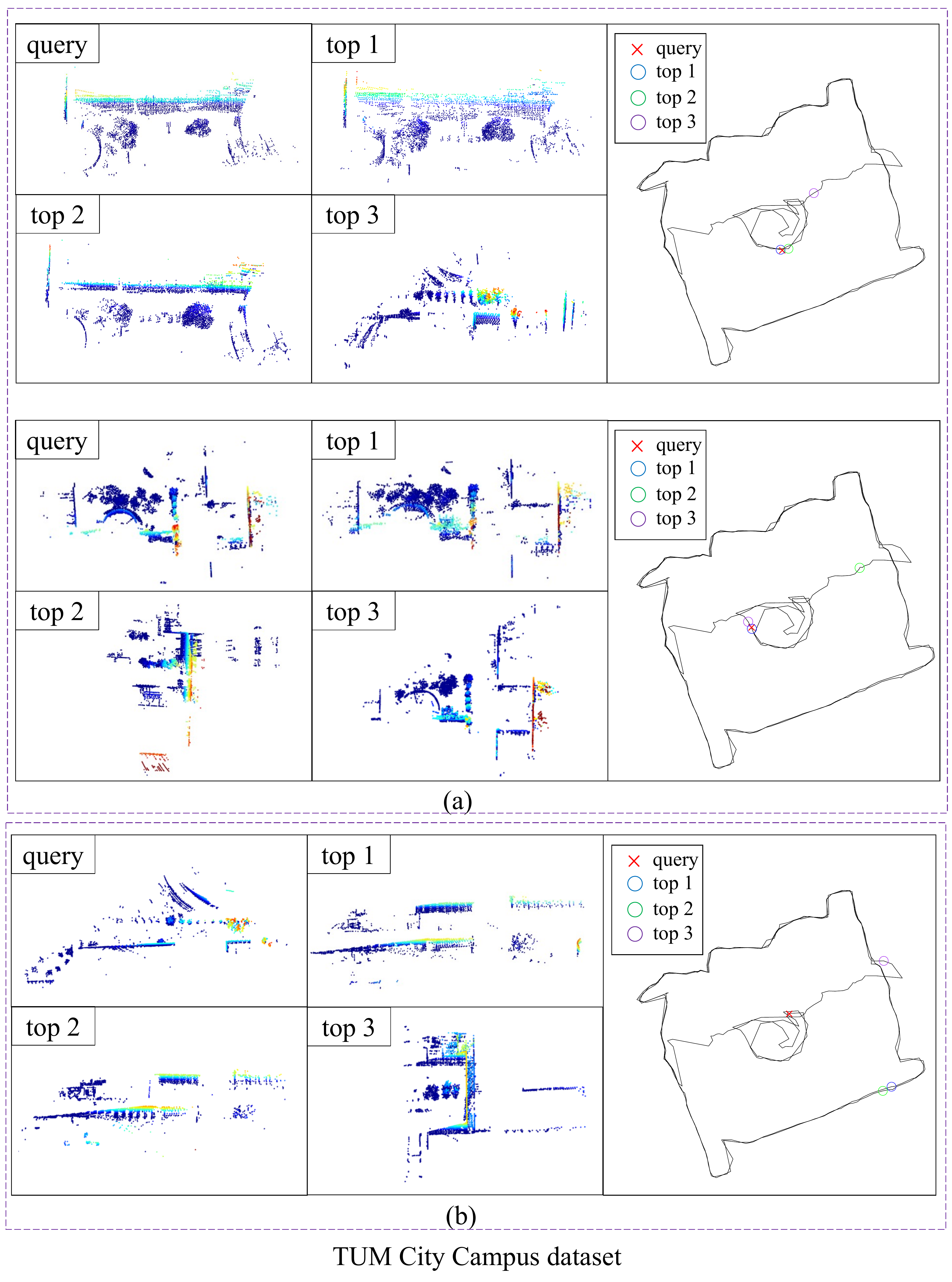}    
\caption{Example retrieval results of CASSPR on the TUM dataset, best viewed on a screen. For each retrieval, the query point cloud and top 3 retrievals are shown. The point clouds' origins are also indicated in a reference map on the right. a) shows the successful retrieval cases, while b) shows a failure case.} 
\label{fig: tum_vis_results} 
\end{figure}

\subsection{Results visualization on the TUM dataset} \label{sec: tum_results_visualization}

In addition to quantitative results, we show qualitative results of two correctly retrieved matches and one failure case in Fig.~\ref{fig: tum_vis_results}. 
A full traversal is made as the reference map on the TUM dataset. 
Then we choose three query point clouds from the traversal, with each representing a single scan from the testing areas. 
For each scan, the query point cloud and the top 3 retrieved matches are shown on the left. 
It is clear that the best match has a very similar geometry to the query point cloud. 
We also display the location of each point cloud in the reference map on the right. 
For each query, the location of the top 1 result (indicated by the blue circle) is correctly overlapped with the query location (represented by the red cross).

\subsection{Evaluation of loop closure detection} \label{sec: loop closure detection}
In this section, we evaluate loop closure detection performance on the KITTI Odometry benchmark~\cite{geiger2012we}, sequences 00 and 08, using the same evaluation protocol as in the previous works~\cite{cattaneo2022lcdnet, chen2021overlapnet, ma2022overlaptransformer, ma2023cvtnet, barros2022attdlnet}. The performance is reported in Table \ref{tab:loop-closure} above.
Note that only values that are available in the original publications are reported. CASSPR achieves the best performance, and only for Average Precision (AP) on the sequence 00, our method is on par with LCDNet~\cite{cattaneo2022lcdnet}.

\begin{table}
\caption{Comparison of loop closure detection performance on the KITTI Odometry benchmark. The sequence number is reported in parentheses, e.g. (00).}\label{tab:loop-closure}
    \centering
    \scalebox{0.76}{
    \begin{tabular}{ccccc}
    \hline
    & AP (00) & AR@1 (00) & AUC (00) & F1 (08)\\
    \hline
    LCDNet~\cite{cattaneo2022lcdnet} & \textbf{0.97} & - & - & - \\
    OverlapNet~\cite{chen2021overlapnet} & - &  81.6 & 86.7 & - \\
    OverlapTransfomer~\cite{ma2022overlaptransformer} & - &  90.6 & 90.7 & - \\
    CVTNet~\cite{ma2023cvtnet} & - & - & 91.1 & - \\
    AttDLNet~\cite{barros2022attdlnet} & - & - & - & 0.13\\
    \hline
    CASSPR (Ours) & 0.96 & \textbf{97.9} & \textbf{92.9} & \textbf{0.30} \\
    \hline
    \end{tabular}
}
\end{table}

\subsection{Number of lightweight self-attention units}
In this section, we explore the network performance with different numbers of lightweight self-attention (LSA) units on the Oxford RobotCar~\cite{maddern20171} and in-house datasets. Specifically, we insert LSA units one by one after each convolutional layer in CASSPR. The network is denoted as CASSPR\_LSA. '0' means that we do not add any LSA units.

Table~\ref{table:number of LSA} shows results of average recall at top 1\% and top 1 with different numbers of LSA units for the CASSPR\_LSA architecture. 
As seen from the table, CASSPR\_LSA achieves the best performance with 6 LSA units. This implies the network has the best global awareness when all attention units are used.

\begin{table*}[ht]
\centering
\caption{Average recall ($\%$) at top 1$\%$ (@1$\%$) and top 1 (@1) for CASSPR\_LSA with different numbers of LSA units trained only on the Oxford RobotCar.}
\label{table:number of LSA}
\begin{tabular}{ccccccccc} 
\hline
\multirow{2}{*}{Number of LSA} & \multicolumn{2}{c}{Oxford RobotCar}   & \multicolumn{2}{c}{U.S.}      & \multicolumn{2}{c}{R.A.}      & \multicolumn{2}{c}{B.D.}       \\
                               & AR @1~        & AR @1\%       & AR @1~        & AR @1\%       & AR @1         & AR @1\%       & AR @1         & AR @1\%        \\ 
\hline
0                              & 93.0          & 97.9          & 86.7          & 95.0          & 80.4          & 91.2          & 81.5          & 88.5           \\
1                             & 93.5          & 98.0          & 86.9          & 95.0          & 80.0          & 88.8          & 80.9          & 87.3           \\
2                             & 93.5          & 98.0          & 87.2          & 96.0          & 80.2          & 89.3          & 80.8          & 87.4           \\
3                              & 92.9          & 97.6          & 90.1          & 96.5          & 81.0          & 91.1          & 82.6          & 89.2           \\
4                             & 92.9          & 97.6          & 87.2          & 95.6          & 82.7          & 92.5          & 82.5          & 89.1           \\
5                              & \textbf{94.7} & \textbf{98.4} & 88.0          & 94.4          & 86.2          & 93.4          & 82.7          & 89.0           \\
6                              & \textbf{94.7} & \textbf{98.4} & \textbf{91.4} & \textbf{97.1} & \textbf{86.8} & \textbf{93.5} & \textbf{86.3} & \textbf{91.0}  \\
\hline
\end{tabular}
\end{table*}

\subsection{The maximum range of 3D LiDAR scans}
\label{sec: maximum range}
\begin{figure}[t!]
\centering
\includegraphics[width=0.47\textwidth]{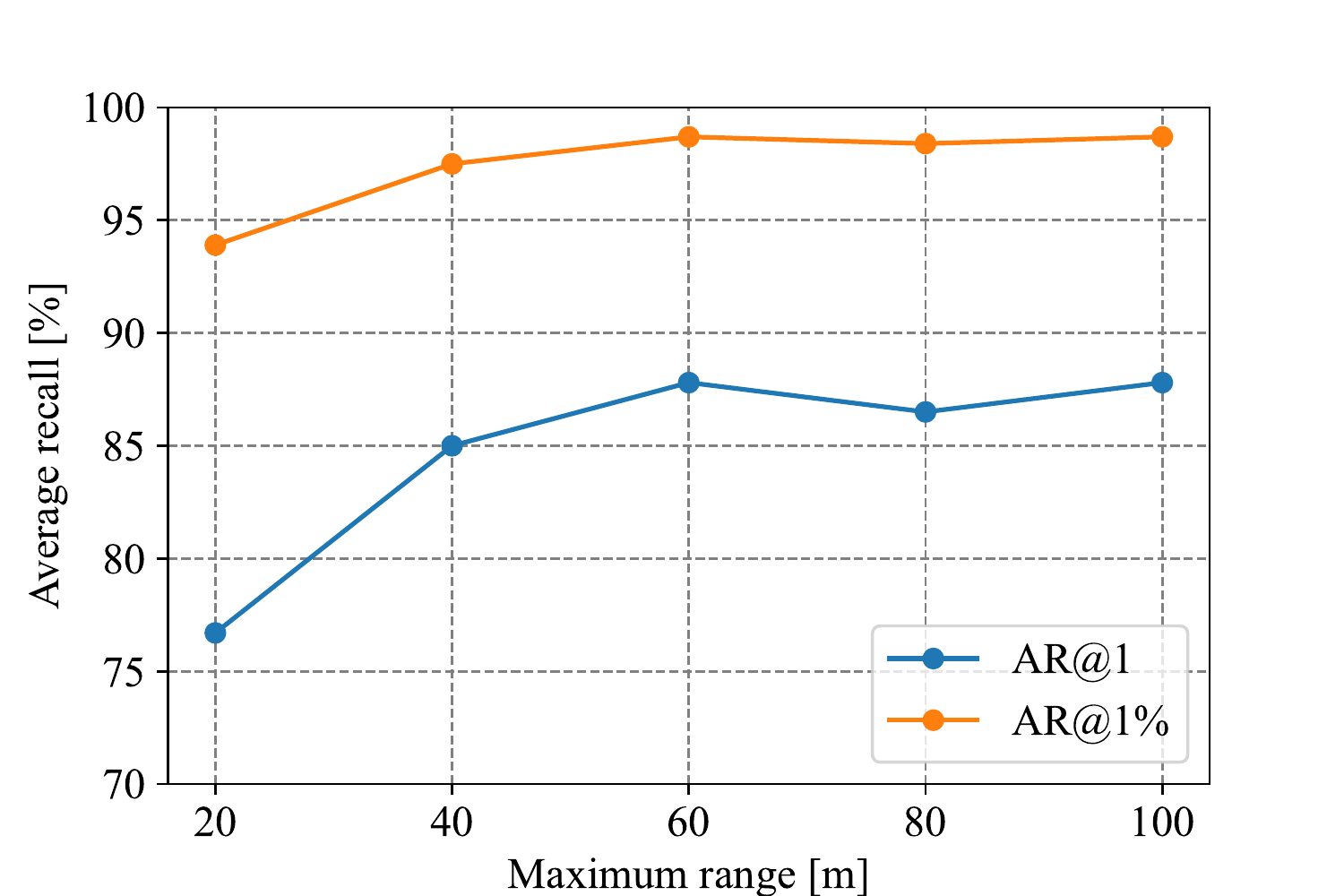}    
\caption[Average recall of CASSPR tested on USyd with a different maximum distance of points]{Average recall of CASSPR tested on the USyd with a different maximum distance of points from the scanner position. Maximum ranges less than \SI{40}{m} show a drop in performance.} 
\label{fig: maximum_range} 
\end{figure}

In this section, we test the performance of the proposed CASSPR with different maximum ranges of points from LiDAR. The experiments are conducted on the USyd dataset~\cite{zhou2020developing}. We set the maximum range of points from \SI{20}{m} to \SI{100}{m} at \SI{20}{m} intervals since the Velodyne VLP-16 sensor utilized in the USyd dataset has a range of about \SI{100}{m}. Fig.~\ref{fig: maximum_range} presents the results, the best performance is obtained when the maximum measurement range is set to at least 60m. 

We can get two conclusions:(1) Compared to the performance from \SI{40}{m} to \SI{100}{m},  our method achieves fairly good  average recall at top 1 ranging from 85.0\% to 87.8\%, demonstrating our CASSPR is robust to the different maximum ranges of points.
(2) When the maximum range is \SI{20}{m}, the performance at top 1 drops to 77.7\% due to the limited information provided by the single scan within a small range.

\section{Conclusion}
We proposed CASSPR as a cross attention transformer for single scan based place recognition. In a dual-branch hierarchical cross attention transformer, it combines both the multi-scale spatial context of voxel-based approaches with the local precision of point-based approaches. This way, we  compensate for quantization losses of voxel-based approaches and introduce long-range context dependency. 
Extensive experiments demonstrate that CASSPR improves the retrieval performance over the state-of-the-art significantly.
In particular, the results confirm that CASSPR is robust to quantization losses.
Future work will explore its integration into Simultaneous Localization and Mapping (SLAM) pipelines for efficient temporal aggregation of point clouds and robustness to moving objects.

\vspace{0.5cm}
\noindent
{\bf Acknowledgements.} This work was supported by the ERC Advanced Grant SIMULACRON, by the Munich Center for Machine Learning, by the EPSRC Programme Grant VisualAI EP/T028572/1, and by the Royal Academy of Engineering (RF\textbackslash 201819\textbackslash 18\textbackslash 163).

{\small
\bibliographystyle{ieee_fullname}
\bibliography{egbib}
}

\end{document}